\documentclass{article}
\usepackage{PRIMEarxiv}
\usepackage[utf8]{inputenc} 
\usepackage[T1]{fontenc}    
\usepackage{hyperref}       
\usepackage{url}            
\usepackage{booktabs}       
\usepackage{amsfonts}       
\usepackage{nicefrac}       
\usepackage{microtype}      
\usepackage{lipsum}
\usepackage{fancyhdr}       
\usepackage{graphicx}       
\usepackage[title]{appendix}
\usepackage[square,numbers]{natbib}
\usepackage{float}
\usepackage{amsthm}
\usepackage{caption}
\usepackage[framemethod=tikz]{mdframed}
\usepackage{etoolbox} 
\usepackage{orcidlink}
\usepackage{listings}
\usepackage{tabularx}
\usepackage{multirow}
\usepackage{longtable}
\usepackage{booktabs}
\graphicspath{{media/}} 

\theoremstyle{definition}

\newmdtheoremenv[
hidealllines=true,
leftline=true,
innerleftmargin=10pt,
innerrightmargin=10pt,
skipabove=10pt,
skipbelow=10pt,
]{prompt}{Prompt}

\title{Characterizing the Investigative Methods of Fictional Detectives with Large Language Models}

\author{
	Edirlei Soares de Lima \orcidlink{0000-0002-2617-3394}\\
	Academy for AI, Games and Media \\
	Breda University of Applied Sciences \\
	Breda, The Netherlands\\
	\texttt{soaresdelima.e@buas.nl} \\
	\And
	Marco A. Casanova \orcidlink{0000-0003-0765-9636}\\
	Department of Informatics \\
	PUC-Rio \\
	Rio de Janeiro, Brazil\\
	\texttt{casanova@inf.puc-rio.br} \\
	\And
	Bruno Feijó \orcidlink{0000-0003-4441-2632}\\
	Department of Informatics \\
	PUC-Rio \\
	Rio de Janeiro, Brazil\\
	\texttt{bfeijo@inf.puc-rio.br} \\
	\And
	Antonio L. Furtado \orcidlink{0000-0003-3710-624X}\\
	Department of Informatics \\
	PUC-Rio \\
	Rio de Janeiro, Brazil\\
	\texttt{furtado@inf.puc-rio.br} \\
}

\begin{document}
	\maketitle
	
	\begin{abstract}
		Detective fiction, a genre defined by its complex narrative structures and character-driven storytelling, presents unique challenges for computational narratology, a research field focused on integrating literary theory into automated narrative generation. While traditional literary studies have offered deep insights into the methods and archetypes of fictional detectives, these analyses often focus on a limited number of characters and lack the scalability needed for the extraction of unique traits that can be used to guide narrative generation methods. In this paper, we present an AI-driven approach for systematically characterizing the investigative methods of fictional detectives. Our multi-phase workflow explores the capabilities of 15 Large Language Models (LLMs) to extract, synthesize, and validate distinctive investigative traits of fictional detectives. This approach was tested on a diverse set of seven iconic detectives -- Hercule Poirot, Sherlock Holmes, William Murdoch, Columbo, Father Brown, Miss Marple, and Auguste Dupin -- capturing the distinctive investigative styles that define each character. The identified traits were validated against existing literary analyses and further tested in a reverse identification phase, achieving an overall accuracy of 91.43\%, demonstrating the method's effectiveness in capturing the distinctive investigative approaches of each detective. This work contributes to the broader field of computational narratology by providing a scalable framework for character analysis, with potential applications in AI-driven interactive storytelling and automated narrative generation.
	\end{abstract}
	
	
	\section{Introduction}
	
	Detective fiction has long been a significant subgenre within the broader field of crime and mystery literature, characterized by its focus on uncovering hidden truths and reconstructing past events through observation, logical analysis, and reasoning \cite{link2023defining}. The foundational ideas for detective stories can be traced back to early narratives, such as the old Persian tale \textit{The Three Princes of Serendip}, where characters demonstrate an uncanny ability to infer truths from seemingly unrelated observations. This concept, famously described by Horace Walpole in 1754, provided the basis for the term \textit{serendipity}, referring to the ability to make unexpected discoveries by combining sharp observation with keen insight \cite{walpole1754serendipity}. As Walpole explained in his famous letter, the princes of Serendip were ``always making discoveries, by accidents and sagacity, of things which they were not in quest of'', such as deducing that a lost camel was blind in one eye by noting that it only ate grass from one side of the path, leaving intact the better quality grass visible on the other side \cite{friedel2001serendipity}. 
	
	This early form of inferential reasoning later evolved into the foundational methods of modern fictional detectives. Edgar Allan Poe's Auguste Dupin, often credited as the first modern fictional detective, relied on what can be described as \textit{retrograde operations}, a stepwise analytical approach to uncovering hidden truths. Similarly, Arthur Conan Doyle's Sherlock Holmes is famously recognized for his \textit{reasoning backwards} method, emphasizing the importance of reconstructing past events from their observed effects. Both approaches are now recognized as forms of \textit{abductive reasoning}, a type of logical inference formally defined by the semiotician Charles Sanders Peirce and widely regarded as a critical component of criminal investigation \cite{eco1988sign}. This form of reasoning, which seeks the most plausible explanation for a set of observations, became central to the genre, as acknowledged by Arthur Conan Doyle in his creation of Sherlock Holmes \cite{carson2009abduction}. Additionally, the idea that detective stories inherently involve a dual narrative structure, comprising both the story of the crime and the story of the investigation, was further formalized by Tzvetan Todorov, who identified this as a defining feature of the genre \cite{todorov1977poetics}.
	
	While detective fiction has a rich tradition in literary studies, the systematic characterization of detective investigative methods remains a relatively underexplored area within \textit{computational narratology}. This field, which seeks to integrate literary narrative theories into the design of narrative generation systems, emphasizes the importance of capturing both the structural and thematic elements of narratives to enable systems to create more contextually rich stories \cite{cavazza2006narratology}. Within this context, detective stories stand out as a particularly challenging subgenre due to their distinct \textit{epistemic} plot structures. As noted by Ryan \cite{ryan2008interactive}, these narratives challenge both readers and computational systems to look through clues and separate relevant information from misleading distractions, making them ideal candidates to explore narrative generation methods. However, the distinctive investigative methods that define fictional detectives, ranging from deductive logic to abductive reasoning, have not yet been systematically captured in a form that can be directly applied to AI-driven narrative generation systems. 
	
	To bridge this gap, the objective of this work is to systematically characterize the investigative methods of fictional detectives using a multi-large language model (LLM) approach. Our method is designed to extract, synthesize, and validate the distinctive investigative traits that define each detective's approach to solving crimes. In essence, this approach seeks to ``investigate the investigators'', exploring the generative and analytical capabilities of multiple LLMs to capture the core components of investigative methods. For this study, we selected a diverse set of iconic detectives, including Hercule Poirot, Sherlock Holmes, William Murdoch, Columbo, Father Brown, Miss Marple, and Auguste Dupin, each representing different investigative styles and narrative traditions. To characterize their investigative methods, our method employs a multi-phase workflow, including description generation, trait extraction, semantic grouping, consistency analysis, and reverse identification, with each phase integrating insights from multiple LLMs to ensure comprehensive trait identification. The effectiveness of the proposed method was evaluated through a series of experiments, including a final validation that confirmed the alignment of the identified traits with existing literary studies, demonstrating its ability to capture meaningful and distinguishing traits for a set of fictional detectives.
	
	The remainder of the paper is structured as follows. Section~\ref{sec2} covers related work, providing an overview of previous studies on detective archetypes and computational approaches to narrative generation. Section~\ref{sec3} describes the proposed method to characterize the investigative methods of fictional detectives using an LLM-based workflow. Section~\ref{sec4} presents the results of our experimental evaluation, including an analysis of the identified investigative traits and the accuracy of our method. Finally, Section~\ref{sec5} offers concluding remarks.
	
	\section{Related Work}
	\label{sec2}
	
	The theoretical foundation of this work draws heavily on the principles of computational narratology, a transdisciplinary field \cite{feijo2021transdisciplinary} pioneered by Mark Cavazza and David Pizzi \cite{cavazza2006narratology}. In their seminal work, they emphasized the importance of integrating literary narrative theories into the development of narrative generation and interactive storytelling systems. This perspective aligns with Mieke Bal's classic framework for narratology \cite{bal2007narratology}, which distinguishes three fundamental narrative levels: \textit{fabula} (the events being told), \textit{story} (the way those events are structured), and \textit{text} (the final, media-specific representation of the narrative). Early computational approaches, including our initial work on plan-based detective story generation \cite{barbosa2014generation}, primarily focused on the fabula level, treating plots as sequences of predefined events structured through logic programming and plan generation.
	
	Moving beyond the fabula level to address the story level requires a more nuanced understanding of narrative forms, including literary genres. Frye's influential analysis of literary genres \cite{frye2020anatomy} identified four major genres: \textit{comedy}, \textit{romance}, \textit{tragedy}, and \textit{satire}, to which Marie-Laure Ryan later added the \textit{mystery} genre \cite{ryan2008interactive}, characterized by \textit{epistemic plots} driven by the desire to uncover hidden truths. Within the broader mystery genre, detective stories form a distinct sub-genre, defined by the dual narrative structure described by Tzvetan Todorov \cite{todorov1977poetics}, which separates the story of the crime from the story of the investigation. With the advent of LLMs, we have recently expanded our approach to cover the story level more comprehensively, incorporating natural language and broader genre considerations, as demonstrated in our recent work on multi-genre story generation \cite{delima2025multigenre}. This progression reflects a broader trend in computational narratology, moving beyond purely structural plot representations to more nuanced character-driven storytelling \cite{delima2023managing}. Our current study contributes along this direction by systematically identifying and categorizing the investigative traits of fictional detectives, thus providing a foundational framework that can be used for the development of new narrative generation methods within this genre.
	
	In parallel, several literary analyses have focused on characterizing the distinctive investigative methods of fictional detectives, often focusing on archetypes that define the genre. Monico \cite{monico2021agatha} examines the contrasting approaches of Hercule Poirot and Miss Marple, emphasizing how their investigative styles reflect broader cultural and gender dynamics, with Poirot characterized by psychological profiling and logical deduction, and Marple by intuition and social observation. Ketović \cite{ketovic2019detective} categorizes detectives into key archetypes, including private investigators, police detectives, amateur detectives, and more specialized subtypes, each defined by distinct investigative strategies and narrative roles. Domjanović \cite{domjanovic2019comparative,domjanovic2022murder} provides a comparative analysis of Poirot and Marple, highlighting the differences in their investigative methods and the influence of their respective social contexts. Similarly, Jenner \cite{jenner2013follow} contrasts evidence-driven, analytical methods often found in early detective fiction with more subjective, psychologically driven approaches that emerged in recent media, reflecting broader cultural shifts in the portrayal of detective styles. While these works provide valuable insights into the investigative methods of selected detectives, they often focus on a limited set of characters. In contrast, our AI-driven approach offers a scalable framework for systematically characterizing a broader range of fictional detectives, enabling automated analysis and potential expansion to lesser-known characters or to identify emerging trends in the genre.
	
	While the studies mentioned above provide valuable theoretical foundations for understanding detective archetypes and methodologies, recent advancements in AI and LLM-based narrative generation offer a promising support for expanding this analysis. Research in this area has explored various approaches to enhance story generation, including fine-tuning models on large-scale datasets for improved character consistency and plot coherence \cite{fan2018,vartinen24}, reinforcement learning for goal-driven storytelling \cite{alabdulkarim2021}, and external knowledge retrieval for maintaining logical consistency \cite{xu2020}. Other works have focused on refining story structure through techniques such as backward reasoning \cite{castricato21}, iterative planning \cite{xie2024}, and recursive reprompting \cite{yang2022}. These methods primarily focus on the models themselves, often lacking a structured, character-centric framework such as the one proposed in our work. Our approach, which systematically identifies and categorizes investigative traits, can serve as a foundational layer for future narrative generation systems, enabling more authentic character-driven storytelling. Additionally, previous works have demonstrated the potential of reusing narrative patterns \cite{lima2024Pattern,delima2024imagining,limaVideojogos2025,delima2021computational} and semiotic relations \cite{limaSBGames23,limaICEC23,limaENTCOM24} to enhance thematic consistency and coherence in AI-generated stories, suggesting a complementary direction for integrating the detective profiles identified in this study.
	
	\section{Methodology}
	\label{sec3}
	
	The methodology proposed to characterize the investigative methods of fictional detectives is structured around an LLM-based workflow comprising five sequential phases: (1) Description Generation, in which each LLM independently generates concise textual descriptions capturing distinctive investigative traits for each detective; (2) Trait Extraction, where each description is processed by an LLM to extract a structured list of investigative method components; (3) Semantic Grouping, where these components are clustered based on their semantic similarity; (4) Consistency Analysis and Trait Synthesis, which quantitatively evaluates consensus among models, retaining only the most consistently identified traits representative of each detective's investigative methods; and (5) Validation via Reverse Identification, which verifies the distinctiveness of the synthesized traits by requiring LLMs to identify detectives solely from their trait descriptions. Detailed explanations of each phase are provided in the following subsections, beginning with the rationale for selecting the set of fictional detectives used in this study.
	
	\subsection{Detective Selection}
	\label{sec31}
	
	The sample of fictional detectives selected for this study -- Hercule Poirot, Sherlock Holmes, William Murdoch, Columbo, Father Brown, Miss Marple, and Auguste Dupin -- reflects a carefully chosen set of popular and influential figures within the detective fiction genre. This selection was guided primarily by the cultural significance and widespread recognition of these characters, as demonstrated by their prominent roles in both literary works and successful television adaptations. Additionally, this set of detectives provides a balanced representation of the major types of fictional investigators, capturing the diversity of approaches within the genre:
	
	\vspace{5mm}
	\begin{itemize}
		\item \textbf{Private investigators:}
		\begin{itemize}
			\item \textbf{Sherlock Holmes} (Arthur Conan Doyle, 1887; TV series \textit{Sherlock}, 2010) exemplifies the application of logical reasoning over detailed observation. Like the princes of Serendip, he refined his observational acuity through a broad range of disciplines, focusing only on knowledge essential for his role as a ``consulting detective''.
			\item \textbf{Hercule Poirot} (Agatha Christie, 1920; TV series \textit{Agatha Christie's Poirot}, 1989) emphasizes psychological profiling. The long-running TV series starring David Suchet provided memorable adaptations of all the Belgian detective's stories.
		\end{itemize}
		\item \textbf{Amateur detectives:}
		\begin{itemize}
			\item \textbf{Miss Marple} (Agatha Christie, 1927; TV series \textit{Miss Marple}, 1984) relies on her skill to find parallels between human behavior in her small village and current criminal actions. She frequently makes discoveries through gossip, a daily occupation that suits her temperament and sharp observational skills.
			\item \textbf{Father Brown} (G. K. Chesterton, 1910; TV series \textit{Father Brown}, 2013) employs a compassionate and morally intuitive approach, drawing on his religious understanding of sin and repentance to uncover hidden truths. His unique insight allows him to discern the true situation where others might be misled by surface appearances.
			\item \textbf{Auguste Dupin} (Edgar Allan Poe, 1841; TV miniseries \textit{The Fall of the House of Usher}, 2023) relies on ``ratiocination'' -- a purely mental, stepwise analysis that he casually reveals to his interlocutors, often with the aim of provoking admiration. Unlike professional detectives, he solves crimes as a personal intellectual exercise.
		\end{itemize}
		\item \textbf{Police detectives:}
		\begin{itemize}
			\item \textbf{Columbo} (TV series \textit{Columbo}, 1968) relies on persistent questioning, using his unassuming demeanor and feigned incompetence to disarm self-assured criminals, gradually cornering them into revealing their guilt. His approach, inspired by the classical dialogue method of questioning attributed to the philosopher Plato \cite{murray2000classical}, involves leading suspects through a series of seemingly casual questions that gradually lead them to admit their guilt.
			\item \textbf{William Murdoch} (Maureen Jennings, 1997; TV series \textit{Murdoch Mysteries}, 2008) combines early forensic methods with technical innovations. His extensive scientific and technical knowledge, often showcased through his inventive gadgets, earned him the nickname ``the Artful Detective'' in the TV series.
		\end{itemize}
	\end{itemize}
	
	The inclusion of Dupin acknowledges his historical significance as the first classic fictional detective, establishing many of the conventions that would later define the genre. Father Brown, in contrast, introduces a unique perspective shaped by his role as a clergyman, offering insights into human morality and sin that distinguish him from other amateur detectives. William Murdoch represents the scientific spirit of his time, combining early forensic methods with technical innovations. Sherlock Holmes and Hercule Poirot exemplify the logical and psychological approaches, each representing iconic figures known for their distinct investigative styles. Miss Marple adds the perspective of an intuitive, observant amateur detective who relies on her deep understanding of human nature, while Columbo demonstrates the effectiveness of persistent questioning and psychological manipulation in extracting confessions.
	
	\subsection{Description Generation}
	\label{sec32}
	
	In the first phase of our LLM-based workflow to characterize the investigative methods of the selected detectives, multiple LLMs are employed to generate concise textual descriptions capturing each detective's distinctive investigative traits. This phase explores the diverse perspectives offered by different LLM architectures and training paradigms, helping minimize model-specific biases or inaccuracies. Specifically, we utilize a set of 15 LLMs from 7 distinct providers (OpenAI, Google, Anthropic, Alibaba, Meta, Mistral AI, and DeepSeek), as detailed in Table~\ref{tab:llm_models}.
	
	\begin{table}[htbp]
		\caption{Overview of the LLMs used in our experiments.}
		\label{tab:llm_models}
		\begin{tabular}{lll}
			\toprule
			\textbf{Model} & \textbf{Description} & \textbf{Version} \\ 
			\midrule
			GPT-3.5 Turbo & \raggedright  Fastest legacy GPT model developed by OpenAI. & gpt-3.5-turbo-0125 \\[2pt]
			GPT-4o & \raggedright  Flexible and high-performing GPT model by OpenAI. & gpt-4o-2024-08-06 \\[2pt]
			GPT-4.1 & \raggedright  Flagship model for complex tasks by OpenAI. & gpt-4.1-2025-04-14 \\[2pt]
			o4-mini & \raggedright  Lightweight reasoning model provided by OpenAI. & o4-mini-2025-04-16 \\[2pt]
			o3-mini & \raggedright  Compact reasoning model offered by OpenAI. & o3-mini-2025-01-31 \\[2pt]
			Gemini 1.5 Pro & \raggedright  Mid-side model offered by Google. & gemini-1.5-pro \\[2pt]
			Gemini 2.0 Flash & \raggedright  Lightweight flagship model provided by Google. & gemini-2.0-flash \\[2pt]
			Claude 3.5 Haiku & \raggedright Fastest model offered by Anthropic. & claude-3-5-haiku-20241022 \\[2pt]
			Claude 3.7 Sonnet & \raggedright  Flagship model offered by Anthropic. & claude-3-7-sonnet-20250219 \\[2pt]
			Llama 3.3 & \raggedright  70-billion-parameter model by Meta. & llama3.3:70b-instruct-q5\_K\_M \\[2pt]
			Qwen 2.5 & \raggedright  72-billion-parameter model created by Alibaba. & qwen2.5:72b-instruct-q5\_K\_M \\[2pt]
			QwQ & \raggedright  32-billion-parameter reasoning model by Alibaba. & qwq:32b \\[2pt]
			Mistral Small & \raggedright  22-billion-parameter model by Mistral AI. & mistral-small:22b-instruct-q5\_K\_M \\[2pt]
			Gemma 3 & \raggedright  27-billion-parameter model developed by Google. & gemma3:27b \\[2pt]
			DeepSeek R1 & \raggedright  70-billion-parameter reasoning model by DeepSeek. & deepseek-r1:70b \\[2pt]
			\bottomrule
		\end{tabular}
	\end{table}
	
	To ensure that the LLMs focused specifically on the task of describing the detectives' investigative approaches, we employed a carefully designed prompt that constrained and guided the generation process. The prompt was crafted to extract descriptions emphasizing only the distinctive methods, reasoning styles, and investigative strategies employed by each fictional detective. It imposed strict structural and stylistic constraints, requiring responses to be limited to a single paragraph of no more than five sentences. This ensured comparability across responses and facilitated the next phases of the workflow. The exact content of the prompt, with the detective's name inserted via the parameter \(D_{name}\), is presented in Prompt~\ref{prompt1}.
	
	\begin{prompt}
		\label{prompt1}
		\
		\textit{Task: Generate a concise and formal description of the distinguishing characteristics of the investigative method used by the fictional detective \(D_{name}\).}
		
		\textit{Requirements:}
		\begin{itemize}
			\item \textit{Base your description of the investigative method used on stories where \(D_{name}\) is the protagonist or a principal investigator.}
			\item \textit{Focus solely on the investigative approach, strategies, and distinguishing features that define this detective's method of solving cases.}
			\item \textit{Consider only the distinguishing characteristics that set the investigative method of \(D_{name}\) apart from those of other fictional detectives.}
			\item \textit{Do not include biographical details, story summaries, or references to specific cases.}
			\item \textit{Structure the response as a single paragraph, without bullet points or numbered lists.}
			\item \textit{Do not include any introductory or concluding sentences outside of the description itself.}
			\item \textit{Limit the response to a maximum of 5 sentences.}
		\end{itemize}
	\end{prompt}
	
	The generation process was fully automated through a Python script that interfaced with both remote APIs (Application Programming Interfaces) and local inference endpoints. Proprietary models, such as those from OpenAI, Google, and Anthropic, were accessed via their respective APIs, while open models were executed locally using a self-hosted Ollama server.\footnote{\href{https://ollama.com/}{https://ollama.com/}} Each model was queried independently using identical parameters to ensure uniform conditions across responses. A temperature setting of 0.0 was applied to reduce variability and promote deterministic outputs. The responses were collected asynchronously and subjected to lightweight post-processing to remove any extraneous formatting artifacts, such as reasoning tags or formatting symbols occasionally returned by certain models. The resulting descriptions were stored in a structured JSON format, associating each detective with the outputs of each LLM.
	
	\subsection{Trait Extraction}
	\label{sec33}
	
	The second phase of our workflow focuses on transforming the unstructured textual descriptions generated in the previous phase into structured sets of investigative traits. While the initial descriptions generated by the LLMs provide rich narrative insights into each detective's approach, their free-form nature presents challenges for systematic comparison and semantic analysis. To address this, we introduced a standardized trait extraction procedure designed to isolate and formalize the key components of each investigative method.
	
	For this phase, a single high-performing LLM (OpenAI's GPT-4o) was used to extract concise and non-redundant lists of traits from each description. A structured prompt was employed to ensure consistency and semantic fidelity across all extractions. The prompt explicitly instructed the model to identify only those traits present in the input text, to avoid paraphrasing or inferential generalization, and to format the output as a bullet-point list. The full content of the prompt is shown in Prompt~\ref{prompt2}, where the parameter \(D_{description}\) was dynamically replaced by each LLM-generated description.
	
	\begin{prompt}
		\label{prompt2}
		\
		\textit{Extract the key traits that describe the investigative method in the following text. List each trait as a separate bullet point. Use a formal and concise style. Do not repeat traits, and do not add information that is not present in the text.}
		
		\textit{Text: \(D_{description}\)}
	\end{prompt}
	
	The entire process was automated using a Python script that iterated over the collection of descriptions for each detective and model. GPT-4o was queried using a temperature setting of 0.0 and the returned responses were parsed using a rule-based post-processing function that extracted each trait into a normalized list. This resulted in a structured dataset in which each detective was associated with one list of investigative method traits per LLM.
	
	\subsection{Semantic Grouping}
	\label{sec34}
	
	The third phase of our workflow focuses on organizing the extracted investigative traits into semantically coherent groups for each detective. While the previous phase yielded structured lists of traits derived from individual LLM outputs, those traits were often phrased differently despite referring to conceptually similar ideas. To address this, we implemented an LLM-based semantic grouping approach, in which traits expressing the same or closely related investigative concepts are consolidated into unified groups. This process enables cross-model synthesis and facilitates the identification of core investigative strategies attributed to each detective.
	
	The semantic grouping was performed using OpenAI's GPT-4o, which received the full set of unique traits extracted for each detective. A grouping prompt was used to instruct the model in identifying and merging traits that conveyed the same or closely related investigative strategies, regardless of wording or lexical variation. For each identified group, the model was required to: (1) provide a concise label summarizing the core idea shared by the grouped traits; and (2) list all original trait phrasings that contributed to the group. The exact structure of the prompt is shown in Prompt~\ref{prompt3}, with the parameter \(D_{traits}\) being dynamically populated with the list of traits extracted in the previous phase.
	
	\begin{prompt}
		\label{prompt3}
		\
		\textit{List of Traits:}
		{\(D_{traits}\)}
		
		\textit{Task: Given the list of traits above describing the investigative methods of a fictional detective, group together all traits that express the same or highly similar idea, regardless of phrasing or wording.}
		
		\textit{For each group, provide:}
		\begin{itemize}
			\item \textit{A description of the core idea of the investigative method, taking into account the grouped traits. Use all relevant distinguishing terms when writing the description.}
			\item \textit{The list of original traits belonging to the group.}
		\end{itemize}
		
		\textit{Present the output as a JSON array in the following format:}
		\begin{lstlisting}[columns=fullflexible]
[
  {
    "label": "Description of the core idea of the investigative method using relevant distinguishing terms",
    "traits": [
        "Original phrasing 1",
        "Original phrasing 2"
    ]
  }
]
		\end{lstlisting}
		
	\end{prompt}
	
	The grouping process was fully automated through a Python script that iterated over the set of unique traits associated with each detective. The prompt was submitted to GPT-4o using a temperature setting of 0.0 to promote deterministic outputs. The returned JSON-formatted responses were parsed to extract semantically grouped clusters, each accompanied by a label summarizing the core investigative idea. Additionally, every trait within each group was annotated with metadata linking it to the original LLMs that had generated it. This enriched representation enabled both the consolidation of semantically similar traits and traceability of model-specific contributions, forming the foundation for the consistency analysis conducted in the next phase.
	
	\subsection{Consistency Analysis and Trait Synthesis}
	\label{sec35}
	
	The fourth phase of our LLM-based workflow involves identifying the most noticeable investigative traits for each detective through a model-wise consistency analysis. While the semantic grouping phase produced coherent clusters of conceptually similar traits, the extent to which these traits were independently recognized across multiple models varies. To isolate the most reliable and representative traits, we performed a quantitative evaluation of cross-model consensus for each semantic group.
	
	For each detective, we calculated the proportion of LLMs that contributed at least one trait to each group. This proportion was treated as a consensus score, representing the degree of agreement among models regarding the core investigative characteristic described by that group. Traits were retained only if their associated group surpassed a predefined consistency threshold of 20\% (i.e., traits that were supported by at least 3 LLMs), ensuring that only those traits identified by multiple independent models were included in the final profile of each detective. This approach filters out erroneous traits introduced by single models while preserving features supported by a broader range of LLMs.
	
	The consistency analysis was implemented using a Python script that iterated over the semantic groups of each detective (generated in the third phase of our workflow). Each group was scored based on the number of contributing LLMs relative to the total number of models evaluated for that detective. Groups meeting the threshold were categorized as consistent and retained for synthesis, while the remaining groups were marked as inconsistent and excluded from subsequent phases. The resulting consistent traits form the final synthesized profile of each detective's investigative method, which is then used as input for the validation phase of our workflow.
	
	\subsection{Validation via Reverse Identification}
	\label{sec36}
	
	The final phase of our LLM-based workflow aims to validate the distinctiveness and semantic adequacy of the synthesized traits produced in the previous phase. Specifically, we test whether LLMs can correctly identify each detective based solely on their extracted investigative traits. This reverse identification task serves as an indirect evaluation of the clarity, uniqueness, and informativeness of the synthesized profiles. If LLMs consistently recognize the correct detective given only a list of traits, it provides strong evidence that the traits capture semantically meaningful and distinguishable investigative patterns.
	
	To conduct this validation, we employed a prompt designed to identify the name of a fictional detective based solely on their investigative traits. The prompt explicitly requested the model to select the most likely detective from a list containing the names of the seven detectives used in our study. No additional narrative or biographical context was provided -- only the synthesized list of traits representing each detective's investigative method. This list was formatted as a sequence of bullet points and dynamically inserted into the prompt using the parameter \(D_{profile}\), as shown in Prompt~\ref{prompt4}.
	
	\begin{prompt}
		\label{prompt4}
		\
		\textit{List of Traits:}
		{\(D_{profile}\)}
		
		\textit{You are given a list of traits describing the investigative method of a fictional detective.
			Your task is to identify which detective this description most likely refers to.}
		
		\textit{Choose only from the following list:}
		\begin{itemize}
			\item \textit{Hercule Poirot}
			\item \textit{Sherlock Holmes}
			\item \textit{William Murdoch}
			\item \textit{Columbo}
			\item \textit{Father Brown}
			\item \textit{Miss Marple}
			\item \textit{Auguste Dupin}
		\end{itemize}
		
		\textit{Respond only with the name of the detective, without explanations or additional text.}
	\end{prompt}
	
	The reverse identification task was performed using the same set of 15 LLMs employed in earlier phases (see Table~\ref{tab:llm_models}). For each synthesized profile, all models were independently prompted to identify the corresponding detective. To ensure deterministic behavior and reduce variability, all LLMs were queried using a temperature setting of 0.0. Model predictions were compared to the ground-truth detective associated with each trait list, and evaluation results were computed using two metrics: per-detective accuracy and overall accuracy across all predictions. Additionally, a confusion matrix was generated to examine misclassification patterns and to assess the distinctiveness of each detective's synthesized investigative profile.
	
	\section{Results and Discussion}
	\label{sec4}
	
	\subsection{Analysis of the Identified Investigative Traits}
	
	We start the analysis of the results by examining the final set of consistent investigative traits identified for each detective. These traits, derived from multi-model consensus, represent the most relevant and consistently recognized aspects of each detective's investigative approach. The complete list of identified investigative traits is presented in Appendix~\ref{appendixA}.
	
	Among the seven detectives analyzed in this study, Hercule Poirot stands out for his consistent reliance on psychological insight and methodical reasoning. The most frequently identified traits include his deep understanding of human nature, meticulous observation of behavior and detail, and a systematic and structured approach to investigation. These results align closely with previous scholarly characterizations of Poirot's methodology, which emphasize that his process is deeply rooted in psychological profiling and an unwavering adherence to ``order and method'' \cite{monico2021agatha,domjanovic2022murder}. Rather than relying solely on physical evidence, Poirot prefers to engage what he famously calls his ``little grey cells'' to reconstruct events through mental synthesis and deductive logic, a characteristic consistently highlighted in both fictional and critical accounts \cite{domjanovic2022murder}. Furthermore, his skillful use of conversation to elicit information and his flair for dramatic confrontations are well-documented hallmarks of his investigative style \cite{monico2021agatha}. These findings reinforce the notion that Poirot's effectiveness stems not only from his intellect but also from his perceptive understanding of psychological nuance and human motivation, validating claims that he views all crimes fundamentally as psychological phenomena \cite{bargainnier1980gentle,havlickova2005agatha}.
	
	The profile generated for Sherlock Holmes reflects his widely recognized identity as the stereotypical deduction-based detective. The traits most consistently identified by the LLMs include his extraordinary capacity to observe minute details and derive complex conclusions, his systematic application of logical reasoning and deduction, and his analytical approach supported by scientific principles and empirical validation. These findings align with numerous scholarly interpretations of Holmes's method. Rapezzi et al. \cite{rapezzi2005white} describe his process as closely resembling clinical diagnosis, reliant on deterministic interpretation of signs and clues, while Anderson et al. \cite{anderson2005analysis} argue that Holmes's methodology is more accurately characterized by abduction -- drawing plausible conclusions from incomplete data. His reliance on observation is remarkable, and his skill in deriving conclusions from physical traces has been consistently emphasized in critical literature, including the classic example of deducing travel details from mud spatters in ``The Adventure of the Speckled Band'' \cite{laven2013detection}. Although he occasionally incorporates psychological profiling and understanding of human behavior, Holmes's investigative identity is more dominantly rooted in physical evidence, forensic detail, and disciplined inference. Furthermore, his use of experimentation, disguises, and narrative synthesis illustrates his inventive yet methodical thinking, reinforcing his place as a functionary of rational inquiry and disciplinary control\cite{michalowicz2023magnificent}.
	
	William Murdoch's investigative profile reflects his reputation as a pioneering detective who embraces scientific methods and technological innovation, aligning closely with the traits identified through our LLM-based analysis. The most prominent traits in his profile include a meticulous and scientific approach to evidence collection, the use of early forensic techniques, and a deep commitment to experimental methods and empirical reasoning. These findings align closely with scholarly interpretations of Murdoch's character as a ``Canadian Sherlock Holmes'' \cite{oklopcic2023image}, known for his reliance on rational deduction and scientific rigor. Additionally, Murdoch is portrayed as a figure constantly at the forefront of scientific advancement, often employing new inventions and experimental technologies, including early CCTV prototypes and GPS devices \cite{sheley2020criminal}. This inventive spirit is captured in traits such as his use of experimental methods and openness to new ideas, distinguishing him from more conventionally minded contemporaries. This technical and methodical approach to crime-solving, combined with his meticulous documentation and reliance on empirical evidence, further emphasizes his role as a forward-thinking investigator who challenges traditional policing methods \cite{petrescu2022stylistic}.
	
	Columbo's investigative style is characterized by a unique blend of psychological manipulation, meticulous observation, and relentless questioning, aligning closely with the traits identified through our LLM-based analysis. His most prominent traits include the strategic use of psychological manipulation, characterized by his disarming, unassuming demeanor, and the deliberate feigning of forgetfulness or confusion to extract critical information from suspects. These traits align with Berzsenyi's analysis \cite{berzsenyi2021columbo}, which highlights Columbo's rhetorical inquiry technique, where he uses a sequence of seemingly innocuous questions to lull suspects into a false sense of security while subtly uncovering inconsistencies in their stories. Additionally, Columbo's reliance on subtle psychological tactics and the use of intuition to confirm his initial suspicious is well-documented in analyses of his methods, where he is noted for often identifying the correct suspect early in the investigation, using persistent questioning to verify his theories rather than relying solely on physical evidence \cite{jenner2013follow}. This approach is further supported by Berzsenyi's \cite{berzsenyi2021columbomethod} observations that Columbo's investigations often focus on piecing together seemingly insignificant facts and revealing hidden truths through incremental evidence gathering, reinforcing his reputation as a master of psychological manipulation and patient methodical investigation.
	
	The profile generated for Father Brown captures his distinctive approach as a detective guided by moral intuition and deep psychological insight. The most consistently identified traits include his profound understanding of human nature, intuitive insight into the motives of criminals, and a focus on the why of a crime rather than the how. This aligns with interpretations of Father Brown as a detective who relies on a deep understanding of human frailty and moral character, drawing on his years of experience as a priest to identify the underlying motivations of offenders \cite{ketovic2019detective}. Unlike detectives who focus on physical evidence, Father Brown often anticipates the thoughts and actions of criminals by placing himself in their psychological state, a method described as ``becoming the criminal'' to reconstruct their motivations and understand their moral failures \cite{ketovic2019detective}. This approach emphasizes the recognition of concealed regret, hidden desperation, and moral contradictions, aligning closely with our identified traits, such as empathetic reconstruction of the criminal's moral fall and identification through subtle psychological cues. Furthermore, his approach to crime-solving often involves challenging preconceptions and breaking through the fa\c{c}ades presented by suspects, reflecting his ability to blend into various social settings and draw out confessions through conversation \cite{burns2005rationalism}. According to Burns \cite{burns2005rationalism}, this unique blend of theological insight, psychological acuity, and moral awareness sets Father Brown apart from more empirically focused detectives, highlighting his reliance on moral reasoning over physical proof.
	
	Miss Marple's profile highlights her distinctive approach to detective work, characterized by keen observation, deep understanding of human behavior, and reliance on social dynamics. Central to her method is the ability to gather information through indirect means, such as conversations and social interactions, which allows her to form a comprehensive picture of events without relying on direct confrontation or physical evidence. This aligns closely with existing studies that emphasize her intuitive grasp of human nature and social observation as essential components of her detective method. Bargainnier \cite{bargainnier1980gentle}, for example, describes her approach as relying on analogy, role-playing, and careful observation, combined with a basic distrust of others' statements and a ruthless determination to uncover the truth. This approach, which often involves drawing connections between seemingly unrelated pieces of information, is further supported by analyses highlighting her reliance on memory, intuition, and life experience gathered from her years in the village of St. Mary Mead \cite{domjanovic2019comparative}. These interpretations align closely with the traits identified in our analysis, including her ability to leverage social knowledge, detect subtle patterns in behavior, and use indirect questioning to reveal hidden truths.
	
	The profile generated for Auguste Dupin captures his identity as a detective characterized by meticulous observation, deep psychological insight, and rigorous analytical reasoning (``ratiocination''). This aligns closely with Grella's description of Dupin as combining ``the intuition of the poet with the analytical ability of the mathematician'', attributing his success to this ``dual temperament'' \cite{grella1970murder}. Additionally, Dupin's approach to investigation emphasizes a non-abstract intellectualism and a reliance on reasoning rather than physical force, reflecting his ability to solve puzzles with minimal physical movement and a preference for mental processes over direct confrontation \cite{ketovic2019detective,jenner2013follow}. Furthermore, his distinct psychological complexity, often described as a form of ``psychic duality'', underscores his ability to engage in both creative and meticulous thought, enhancing his capacity to perceive hidden truths and subtle connections \cite{laven2013detection}.
	
	\subsection{Accuracy Analysis}
	
	As the next step in our analysis, we assessed the distinctiveness and reliability of the synthesized trait profiles based on the results of the reverse identification process (Section~\ref{sec36}). In this analysis, we evaluated whether the extracted traits are sufficiently specific to allow the 15 LLMs presented in Section\ref{sec36} to accurately identify each detective based solely on their synthesized profiles.
	
	The results of this evaluation, presented in Table~\ref{tab:detectiveaccuracy}, demonstrate a high overall accuracy of 91.43\%, with 96 out of 105 cases correctly identified. This high accuracy supports the effectiveness of our approach, indicating that the synthesized trait profiles effectively capture the core investigative styles of these detectives. Notably, several detectives, including Hercule Poirot, Columbo, Father Brown, and Miss Marple, achieved perfect scores of 100\%, highlighting the clarity and distinctiveness of their respective profiles.
	
	\begin{table}[htbp] 
		\centering \caption{Accuracy of character identification per detective.} 
		\label{tab:detectiveaccuracy} 
		\begin{tabular}{lc} 
			\toprule 
			\textbf{Detective} & 
			\textbf{Accuracy (Correct / Total)} \\ 
			\midrule 
			Hercule Poirot & 100.0\% (15/15) \\[2pt] 
			Sherlock Holmes & 93.3\% (14/15) \\[2pt] 
			William Murdoch & 93.3\% (14/15) \\[2pt] 
			Columbo & 100.0\% (15/15) \\[2pt] 
			Father Brown & 100.0\% (15/15) \\[2pt] 
			Miss Marple & 100.0\% (15/15) \\[2pt] 
			Auguste Dupin & 53.3\% (8/15) \\[2pt] 
			\bottomrule 
		\end{tabular} 
	\end{table}
	
	To understand the misclassification cases, we decided to examine the error patterns through a confusion matrix (Table~\ref{tab:confusion_matrix}). This analysis revealed that the primary source of error involved the frequent misclassification of Auguste Dupin as Sherlock Holmes. This pattern is consistent with the historical influence that Auguste Dupin, as the first prototype of fictional detective, had on the creation of Sherlock Holmes. It is well-documented that Arthur Conan Doyle drew significant inspiration from Edgar Allan Poe's character Dupin when creating Sherlock Holmes \cite{vanlaethem2017connecting}, including the emphasis on ratiocination, meticulous observation, and analytical reasoning that characterize both detectives. This close conceptual alignment very likely explains the observed pattern of misclassification, as the two characters share a similar investigative style. Additionally, we noted that while the academic literature often identifies abduction as the dominant form of reasoning for both Dupin and Holmes \cite{eco1988sign}, this term, which was present during the trait extraction phase, did not persist through the semantic grouping phase due to the generalization process that combined specific reasoning types under broader categories, such as logical reasoning.
	
	\begin{table}[htbp]
		\caption{Confusion matrix of predicted vs. actual detectives.}
		\label{tab:confusion_matrix}
		\begin{tabular}{lccccccc}
			\toprule
			\textbf{Actual $\backslash$ Predicted} & \textbf{H. Poirot} & \textbf{S. Holmes} & \textbf{W. Murdoch} & \textbf{Columbo} & \textbf{F. Brown} & \textbf{M. Marple} & \textbf{A. Dupin} \\
			\midrule
			Hercule Poirot   & 15 & 0  & 0  & 0  & 0  & 0  & 0  \\[2pt]
			Sherlock Holmes  & 0  & 14 & 1  & 0  & 0  & 0  & 0  \\[2pt]
			William Murdoch  & 0  & 1  & 14 & 0  & 0  & 0  & 0  \\[2pt]
			Columbo          & 0  & 0  & 0  & 15 & 0  & 0  & 0  \\[2pt]
			Father Brown     & 0  & 0  & 0  & 0  & 15 & 0  & 0  \\[2pt]
			Miss Marple      & 0  & 0  & 0  & 0  & 0  & 15 & 0  \\[2pt]
			Auguste Dupin    & 0  & 7  & 0  & 0  & 0  & 0  & 8  \\[2pt]
			\bottomrule
		\end{tabular}
	\end{table}
	
	Interestingly, the misclassification pattern of Auguste Dupin as Sherlock Holmes varied significantly across models. Reasoning-focused LLMs (models designed to simulate a structured thought process), such as o3-mini, o4-mini, DeepSeek R1, and Claude 3.7 Sonnet, correctly identified Dupin, potentially reflecting their capacity to interpret nuanced trait sets with greater contextual awareness. However, QwQ, which is also a reasoning model, failed to differentiate Dupin from Holmes, highlighting that reasoning ability alone is not a guarantee of accurate classification. This inconsistency suggests that while reasoning capabilities can improve the ability to capture subtle distinctions, other factors, such as the model architecture and training data, also play a critical role.
	
	Beyond the misclassification of Auguste Dupin, we identified two isolated cases of misclassification involving other detectives. Mistral Small, the smallest model in our experiment (22 billion parameters), incorrectly identified William Murdoch as Sherlock Holmes. This mistake may be partially attributed to the limited parameter size of the model, potentially reducing its ability to differentiate closely related investigative styles, especially considering that all other models were able to correctly identify William Murdoch.
	
	The second case involved DeepSeek R1, which incorrectly classified Sherlock Holmes as William Murdoch. A review of the reasoning output from DeepSeek R1 revealed that the model carefully weighed the scientific and empirical aspects of each detective's method before settling on a final classification. In its internal analysis, the model initially considered multiple candidates, noting that the description included ``an extraordinary ability to observe minute details and deduce complex conclusions'', a well-known characteristic of Sherlock Holmes, while also recognizing the emphasis on ``scientific methods, experimentation, and empirical evidence'', which it associated more closely with William Murdoch. The model explicitly stated: ``While Sherlock Holmes is known for his logic and deduction, William Murdoch specifically uses more scientific and forensic approaches''. Further into its reasoning, DeepSeek R1 continued to evaluate the role of scientific principles, stating: ``Utilization of scientific knowledge and principles... points towards William Murdoch because he's a pioneer in using forensic methods. Sherlock Holmes does use some forensics, but maybe not as much as Murdoch''. This focus on forensic innovation led the model to prioritize Murdoch, even while acknowledging that traits such as ``systematic and analytical approach'' and ``elimination of impossibilities to determine the truth'' are more traditionally associated with Holmes. Ultimately, the model's decision was influenced by the strong emphasis on scientific methods and technical rigor, concluding: ``Putting it all together... I'm leaning towards William Murdoch because of the focus on scientific knowledge, experimentation, and empirical evidence''.
	
	\section{Concluding Remarks}
	\label{sec5}
	
	The present paper, drawing on concepts from computational narratology research, aimed to systematically characterize the distinct investigative styles of seven fictional detectives. This was achieved through an automated AI-driven process that considered the stories in which these detectives figure as protagonists, compiling detailed lists of traits intended to uniquely identify each of these criminal experts. The detectives were chosen in view of their recognized skills and popularity. To perform the analysis, we employed fifteen LLMs, acting as a well-selected jury, whose members are called to express their perspectives and then strive to conciliate their different interpretations. The decision to employ a plurality of LLMs was motivated by the desire to balance reasoning approaches and mitigate the risk of model-specific biases or hallucinations. As part of their training, the LLMs have acquired varying degrees of familiarity with detective narratives, enabling them to participate in this form of automated literary analysis.
	
	To avoid overreliance on AI-generated outputs, we carefully validated the automatically extracted traits by comparing them against established literary studies, which provided strong evidence that the resulting traits effectively captured the distinctive investigative styles of each of the seven detectives. We then proceeded in reverse direction to verify the sufficiency of the traits to uniquely identify the investigators from whom they had been extracted. This validation also proved successful, achieving a high overall accuracy across the fifteen LLMs, thereby confirming both the robustness of our method and the appropriateness of our selected detective sample. Instances of misclassification were analyzed in detail, revealing that these discrepancies often stemmed from differences in model design or well-known similarities between fictional detectives.
	
	To ensure the reproducibility of our work, we have provided the full set of prompts used to guide the LLMs, which form a critical component of our methodology. These prompts were designed to be flexible, allowing the inclusion of a wide range of fictional detectives without modification. This flexibility makes it possible to extend the approach to different sets of investigators, including those from more contemporary narratives or those characterized by unconventional methods. Additionally, our approach could be adapted to explore other narrative roles, such as criminals or victims, or even to delve into other literary genres, providing a broader foundation for computational narratology research.
	
	Looking ahead, this work also serves as a foundational step toward our next objective: the development of interactive storytelling tools for the generation of detective stories. Beyond its primary goal of systematically characterizing investigative methods, the trait profiles identified in this paper are intended to support the future development of AI-driven systems capable of generating narratives where fictional detectives exhibit contextually accurate investigative behaviors. Specifically, these profiles can guide character actions \cite{delima2023managing}, support the development of branching storylines \cite{delima2022procedural}, and ensure consistency in the portrayal of each detective's unique investigative style, thereby enhancing the depth and authenticity of AI-generated detective stories.
	
	Lastly, we believe that the multi-phase approach presented in this paper, which leverages the collective insights of multiple LLMs, represents a valuable contribution to the emerging field of LLM ensemble methods \cite{lu2024merge}. This flexible workflow, capable of integrating diverse LLM perspectives, is not limited to the analysis of detective traits but can be adapted to other domains where nuanced character or entity profiling is required. As LLMs continue to evolve, we expect this approach to remain relevant, providing a robust framework for future applications that require comprehensive trait identification and contextual analysis.
	
	\bibliographystyle{abbrvnat}
	\bibliography{references}
	
	\vspace{4mm}
	\begin{appendices}
		\section{List of Investigative Traits Identified for Each Detective}
		\label{appendixA}
		
		\begin{longtable}{@{} p{2.3cm} p{11.7cm} r @{}}
			\caption{Consistently identified investigative traits for each detective, with scores indicating the percentage of LLMs that independently supported each trait.}
			\label{tab:trait_consensus} \\
			\toprule
			\textbf{Detective} & \textbf{Trait Description} & \textbf{Score (\%)} \\
			\midrule
			\endfirsthead
			\bottomrule
			\endlastfoot
			\multirow{11}{*}{Hercule Poirot} & Reliance on psychological insight and understanding of human nature & 73.3 \\
			& Meticulous attention to detail and observation of human behavior and psychology & 66.6 \\
			& Systematic and methodical approach to gathering and analyzing information & 66.6 \\
			& Emphasis on logical deduction and reasoning & 46.6 \\
			& Reconstruction of events through logical inferences and mental synthesis & 40.0 \\
			& Use of conversation and interviews to extract information & 33.3 \\
			& Focus on motive and psychological underpinnings & 26.6 \\
			& Use of intuition and insight to identify inconsistencies and hidden clues & 26.6 \\
			& Emphasis on order, symmetry, and method & 26.6 \\
			& Use of theatrical flair and dramatic confrontations & 26.6 \\
			& Presentation of a comprehensive and internally consistent solution & 26.6 \\
			\midrule
			\multirow{13}{*}{Sherlock Holmes} & Extraordinary ability to observe minute details and deduce complex conclusions & 60.0 \\
			& Application of logical reasoning and deduction to solve mysteries & 53.3 \\
			& Systematic and analytical approach to investigation & 46.6 \\
			& Utilization of scientific knowledge and principles in investigation & 40.0 \\
			& Extensive knowledge across various disciplines & 40.0 \\
			& Reconstruction of events through physical clues and mental simulation & 40.0 \\
			& High degree of experimentation and verification & 33.3 \\
			& Psychological profiling and understanding of human behavior & 33.3 \\
			& Use of forensic science and empirical evidence & 33.3 \\
			& Use of creative thinking and unconventional methods & 26.6 \\
			& Use of disguises and undercover strategies for information gathering & 26.6 \\
			& Synthesis of disparate facts into coherent narratives & 26.6 \\
			& Elimination of impossibilities to determine the truth & 26.6 \\
			\midrule
			\multirow{12}{*}{William Murdoch} & Meticulous and scientific approach to evidence collection and analysis & 60.0 \\
			& Use of forensic science and innovative techniques & 60.0 \\
			& Application of experimental methods and technologies & 53.3 \\
			& Rigorous observation, deductive reasoning, and logical analysis & 46.6 \\
			& Use of early forensic techniques such as fingerprinting and blood analysis & 46.6 \\
			& Reconstruction of crimes and crime scenes & 40.0 \\
			& Understanding of human psychology and behavior & 40.0 \\
			& Empirical observation and evidence-based reasoning & 40.0 \\
			& Openness to new ideas and challenging conventional wisdom & 33.3 \\
			& Identification of subtle patterns and connections & 26.6 \\
			& Collaboration with experts from various fields & 26.6 \\
			& Use of photography and documentation techniques & 26.6 \\
			\midrule
			\multirow{13}{*}{Columbo} & Use of psychological manipulation and understanding of human psychology & 80.0 \\
			& Noting and piecing together insignificant facts and inconsistencies & 73.3 \\
			& Disarming and unassuming demeanor to lull suspects into a false sense of security & 60.0 \\
			& Uncovering contradictions and revealing hidden truths through questioning & 53.3 \\
			& Meticulous attention to detail and acute observational skills & 46.6 \\
			& Combination of deceptive demeanor with rigorous observation and analytical skills & 46.6 \\
			& Feigning forgetfulness or confusion as a strategic tool to manipulate interactions & 40.0 \\
			& Revisiting and reexamining evidence and crime scenes & 33.3 \\
			& Strategy of persistent questioning and returning to suspects with additional queries & 33.3 \\
			& Building rapport with witnesses and suspects through casual conversation & 33.3 \\
			& Use of intuition and patience over rapid breakthroughs & 26.6 \\
			& Meticulous reconstruction of the crime scene through incremental evidence gathering & 20.0 \\
			& Securing confessions through the suspect's own words & 20.0 \\
			\midrule
			\multirow{21}{*}{Father Brown} & Profound intuition and deep understanding of human nature & 46.6 \\
			& Emphasis on intuitive insight and psychological understanding & 40.0 \\
			& Observation of subtleties of human behavior and nuances of moral character & 40.0 \\
			& Uncovering motives and intentions that elude conventional detectives & 40.0 \\
			& Ability to identify and challenge preconceptions & 40.0 \\
			& Combination of theology and philosophy with practical knowledge of human frailty & 40.0 \\
			& Ability to blend into various social settings & 40.0 \\
			& Focus on the *why* of a crime rather than the *how* & 40.0 \\
			& Humble and unassuming demeanor & 26.6 \\
			& Anticipating actions and thoughts of criminals & 26.6 \\
			& Solving crimes by understanding the *type* of sinner rather than assembling physical proof & 26.6 \\
			& Distinction from empirically focused detective methods through psychological acuity and moral understanding & 26.6 \\
			& Distinctive blend of psychological acuity, moral awareness, and intellectual curiosity & 20.0 \\
			& Subtle grasp of moral and spiritual insights & 20.0 \\
			& Drawing out confessions or revelations through conversation & 20.0 \\
			& Presumption of inherent goodness in individuals & 20.0 \\
			& Empathetic reconstruction of the criminal's moral fall & 20.0 \\
			& Identification of perpetrators through recognition of concealed regret or desperation & 20.0 \\
			& Use of humility and empathy & 20.0 \\
			& Emphasis on understanding motivations and emotional states of individuals & 20.0 \\
			& Focuses on discerning motives and moral weaknesses & 20.0 \\
			\midrule
			\multirow{11}{*}{Miss Marple} & Reliance on keen observation and acute understanding of human nature and behavior & 53.3 \\
			& Use of psychological insights and understanding of social dynamics & 53.3 \\
			& Utilization of extensive knowledge of village life and its inhabitants & 40.0 \\
			& Gathering information through indirect methods such as conversations and social interactions & 40.0 \\
			& Making connections between past experiences and current events & 40.0 \\
			& Construction of a comprehensive picture of circumstances through observation, deduction, and subtle manipulation of conversations & 33.3 \\
			& Assessment of individuals' credibility and motives without direct questioning or physical evidence & 26.6 \\
			& Use of age and apparent innocence to disarm suspects and gather information & 26.6 \\
			& Listening attentively to gossip and trivial details to uncover hidden connections & 26.6 \\
			& Emphasis on patience, empathy, and intuitive understanding of human frailties & 26.6 \\
			& Leveraging memory and understanding of human nature to identify patterns and motives & 20.0 \\
			\midrule
			\multirow{7}{*}{Auguste Dupin} & Understanding and anticipation of human psychology, behavior, and motivations & 80.0 \\
			& Rigorous and analytical reasoning through observation, deduction, and logical analysis (ratiocination) & 66.6 \\
			& Meticulous observation and attention to detail, noticing subtle cues & 60.0 \\
			& Use of imagination and creative thinking to explore unconventional solutions & 40.0 \\
			& Prioritization of mental processes and intellectual analysis over physical evidence & 40.0 \\
			& Reconstruction of events and uncovering hidden evidence through logical deduction & 33.3 \\
			& Recognition of solutions concealed in plain sight and challenge of conventional assumptions & 20.0 \\
			\addlinespace
			\bottomrule
		\end{longtable}
		
	\end{appendices} 
	
\end{document}